\documentclass[1p]{elsarticle}
\usepackage{lineno,hyperref}
\usepackage{url}
\usepackage{amsmath}
\usepackage[utf8]{inputenc}

\modulolinenumbers[5]

\newcommand{\ff}[1] {  \mbox{\footnotesize{#1}}  }

\journal{Cognitive Systems Research}

\begin{document}

\begin{frontmatter}

\title{Applying Deutsch’s concept of good explanations to artificial intelligence and neuroscience - an initial exploration}
\author{Daniel C. Elton$^a$}
\ead{daniel.elton@nih.gov}
\address[1]{Imaging Biomarkers and Computer-Aided Diagnosis Laboratory, Radiology and Imaging Sciences, National Institutes of Health Clinical Center, Bethesda, MD 20892, USA}

\begin{abstract}
Artificial intelligence has made great strides since the deep learning revolution, but AI systems still struggle to extrapolate outside of their training data and adapt to new situations. For inspiration we look to the domain of science, where scientists have been able to develop theories which show remarkable ability to extrapolate and sometimes predict the existence of phenomena which have never been observed before. According to David Deutsch, this type of extrapolation, which he calls “reach”, is due to scientific theories being hard to vary. In this work we investigate Deutsch’s hard-to-vary principle and how it relates to more formalized principles in deep learning such as the bias-variance trade-off and Occam’s razor. We distinguish internal variability, how much a model/theory can be varied internally while still yielding the same predictions, with external variability, which is how much a model must be varied to accurately predict new, out-of-distribution data. We discuss how to measure internal variability using the size of the Rashomon set and how to measure external variability using Kolmogorov complexity. We explore what role hard-to-vary explanations play in intelligence by looking at the human brain and distinguish two learning systems in the brain. The first system operates similar to deep learning and likely underlies most of perception and motor control while the second is a more creative system capable of generating hard-to-vary explanations of the world. We argue that figuring out how replicate this second system, which is capable of generating hard-to-vary explanations, is a key challenge which needs to be solved in order to realize artificial general intelligence. We make contact with the framework of Popperian epistemology which rejects induction and asserts that knowledge generation is an evolutionary process which proceeds through conjecture and refutation.
 
\end{abstract}

\begin{keyword}
Deep learning, artificial intelligence, intelligence, generalization, extrapolation, interpolation, Occam's razor, simplicity, critical rationalism, induction, Karl Popper
\end{keyword}

\end{frontmatter}


\section{Introduction}
The field of Artificial Intelligence has made great strides since the ascendancy of deep learning since 2012. Deep learning models can now match or exceed human-level performance on natural image classification,\cite{He2015} medical image diagnosis,\cite{Liu2019} and the game of Go.\cite{Silver2016} Several companies have fully autonomous vehicles on the road, and Google's Duplex system has wowed an audience with its ability to engage in natural language conversation. More recently the GPT3 model has demonstrated the able to write very convincing stories and perform tasks during test-time that seem somewhat removed from its training corpus. Yet there are still many things AI cannot do. Today's AI systems lack human-level common sense understanding, are clumsy at the robotic manipulation of objects, and poor at casual reasoning. Another issue is that today's AI cannot learn from a few examples like humans and requires massive amounts of data to train. Most importantly though, today's AI systems are all narrow - they can only perform exactly the tasks they were trained to do, working within the distribution of their training data. As soon as today's AI systems are asked to work outside their training data distribution, they typically fail. Despite these shortcomings, we believe the deep learning paradigm (brute force fitting to large datasets) can go a long way as compute and data become ever more plentiful. However, as we will argue here, certain forms of extrapolation beyond the data fed into the system will always be out of reach to deep learning. The type of extrapolation ability we refer to is best demonstrated by scientific theories. According to David Deutsch, this type of extrapolation is enabled by scientific theories being an example of a larger class of models he calls ``good explanations'' which are defined by the property that they are hard to vary (HTV).\cite{Deutsch2011-DEUTBO} After introducing Deutsch's principle we explore how it might help us better understand human and artificial intelligence. To the author's knowledge this research direction has not been undertaken previously in the published literature. The goal of this paper is to shed light on what the limits of deep learning are and how the HTV principle might be formalized so that it can be measured in AI systems. 

\section{Generalization vs extrapolation}
In our view, the way the term ``generalization'' is used across both statistics and machine learning can often be misleading. In both fields,``generalization ability'' is defined as the gap between test set error and training set error, where the training set and test set are drawn from the same distribution.\cite{hastie01statisticallearning,NeyshaburICLR2020} What we are interested in here is the ability to generalize \textit{across} distributions, not within them, so for sake of clarity we will refer to this as  ``extrapolation''. Deep learning systems are notoriously bad at extrapolation, often failing spectacularly when small distributional shifts occur. To give a few high profile examples, a deep learning system for diagnosing retinopathy developed by Google's Verily Life Sciences which reached human-level diagnostic ability in the lab performed very poorly in field trials in Thailand due to poor lighting conditions, lower resolution images, and images which had been stitched together.\cite{Beede2020} A trained diagnostician would have been able to adapt to these conditions, but deep learning could not. Another high profile example occurred in 2015 when a deep learning system embedded in Google's Photos service was found to be tagging African American people as ``gorillas''. The quick-fix solution was to remove the category of gorillas. Three years later, \textit{WIRED} magazine reported that the category was still missing, suggesting that the problem was not easy to solve.\cite{simonite_2018} In 2017 a much-lauded DeepMind deep reinforcement learning system\cite{Mnih2015} which achieved super-human performance on several Atari games was shown to fail if minor changes are made, such as moving the paddle 3\% higher in the game of Breakout.\cite{Kansky2017} This shows that the system has no grasp of basic physics and is built on top of a highly brittle fitting procedure. Another example is DeepMind's \textit{AlphaStar} system for playing the computer game StarCraft. The system plays at the above-human level for a given character and map but cannot generalize to other characters and maps (it must be retrained separately for each one).\cite{Marcus2018} 

The range of input space within which a model, theory, or explanation makes accurate predictions has different names in different disciplines. The physicist David Deutsch calls it the ``reach'', Popper calls it ``range'',\cite{Popper:LoSD} philosophers sometimes call it the ``explanatory power'', and in some sub-fields of machine learning it is called the ``applicability domain''. As discussed in detail by Hasson et al.,\cite{Hasson2019} and by this author in a prior work,\cite{Elton2020} the double descent phenomena indicates that machine learning models operate largely by local interpolation over a dense set of training data rather than by extracting global trends or general principles. In light of these findings, the failures noted above are not so surprising. 

There are couple of approaches that may help with these failure modes. In a prior work we advocated adding some self-awareness to such systems so they know the bounds of their expertise and can provide a warning to users when they are asked to perform outside their domain of applicability.\cite{Elton2020} Another possible avenue is to develop unsupervised learning algorithms which can adapt on-the-fly as the distribution of their input data changes without forgetting what they've already learned. The influential deep learning scientist Yann Lecun has touted this approach, calling unsupervised learning the key missing ingredient of AI in his famous ``layer-cake'' analogy for AI.\footnote{\href{https://medium.com/syncedreview/yann-lecun-cake-analogy-2-0-a361da560dae}{https://medium.com/syncedreview/yann-lecun-cake-analogy-2-0-a361da560dae}} However, a constant stream of new data is still required for unsupervised learning to work. What interests us here is the ability to generalize beyond training data distributions without having to take in any new data first. A related line of work is ``zero-shot'' learning and more recently, ``less than one shot'' learning,\cite{sucholutsky2020less} which might be thought of as a type of dynamic clustering where spaces between clusters are left as a sort of terra incognita where additional clusters may appear. This is an important line of work, but so far such systems have only been demonstrated on carefully curated toy problems and even on such problems perform poorly, with average accuracies typically less than 50\% . 

What is missing from the aforementioned systems? We believe we can gain insight into this question by turning our attention away from the types of statistical modeling done in mathematics and computer science departments to the type of modeling done in science departments, which is based on theories developed using the scientific method. Physicists in particular have an impressive track record of being able to generate models that predict phenomena their creators had never seen or imagined before. For instance, Isaac Newton developed his laws of motion which have been applied successfully to make predictions in many domains Newton had no direct experience in. In 1915 Einstein predicted that given the initial conditions of a light ray traveling close to the sun, the light ray would follow a curved trajectory, a finding that was confirmed experimentally a five years later during a solar eclipse.\cite{1920starlightbent} Many radical and surprising phenomena have been predicted by physicists in advance, before any empirical confirmation - black holes, grRavitational waves, antimatter, the Higgs boson, quantum computation, and metamaterials. With regards to invention and innovation, much can be done with trial and error in the lab, but explanatory theories can be applied to hypothetical configurations of matter which have never been created before and suggest which configurations would be the most useful for some end. Edison developed the light bulb through trial and error, but it is doubtful the first nuclear weapon or quantum computer could have been developed through trial and error alone - deep explanatory theories were needed to guide such advances. To avoid trial-and-error in the lab, many technological artifacts such as novel drugs or spacecraft are first developed \textit{in-silico} using physics-based simulation. Machine learning (ML) based prediction is much more computationally expedient than physics-based simulation so much work has been done to see if ML can replace physics simulation. However, ML  stops performing outside of domain of data it was trained on, and thus fails precisely in the domains where simulations provide the most value (ie going beyond the domain of experimental data collected so far). To give another example, in the course of our prior work on trying to predict the properties of molecules,\cite{Elton2018,barnes2018machine} a system trained on explosive and propellant molecules predicted that a sugar molecule would have a high explosive energy. Similarly, deep generative models based on deep learning fail to yield truly novel designs. 
To give an example, a deep reinforcement learning system for constructing molecules developed by a team from Insilico Medicine and Harvard received much media fanfare after one of the molecules it generated was synthesized and shown to be active against cancer in a mouse model.\cite{Zhavoronkov2019} However, Walters \& Murcko quickly pointed out that the molecule it generated is very similar to two existing drug molecules which were in the system's training database.\cite{Walters2020} In general deep generative models trained on molecules generate highly nonphysical nonsense molecules when pushed to generate molecules outside their training set distribution.\cite{Elton2019} 

It is not our intention to wade into the philosophical debate about what denotes an ``explanation'' versus a ``pure prediction'', although we think there is an important distinction there to be made. In what follows we treat scientific explanations and predictive models as elements of the same class. This may seem odd, but both are scientific explanations and models can be operationalized as functions which take in inputs (observations/data) and output predictions. Our latest physical theories can make accurate predictions for an enormous range of phenomena and are only known to break down in exotic situations where quantum gravity is important. The question we are interested in is why models derived via the scientific method can extrapolate while models derived from the methods of deep learning and statistics cannot. More generally we are interested in if there are any general characteristics of models/theories/explanations which correlate with their reach (how far they can extrapolate). According to David Deutsch, one such characteristic is that models/theories/explanations with large reach are ``hard-to-vary'' (HTV) while those with small reach are ``easy-to-vary''.\cite{Deutsch2011-DEUTBO} If true this is a profound insight, but to our knowledge it has not yet been formalized or applied to artificial intelligence.  

\section{Understanding critical rationalism and the HTV principle}
What makes Deutsch's HTV principle fascinating is that it was invented within the context of critical rationalism, the epistemology of Karl Popper. There is a close connection between epistemology - the philosophical study of knowledge - and the field of artificial intelligence, which studies (among other things) hows to store and utilize knowledge in computers and how to program computers to discover knowledge. Deutsch has argued that a better understanding of Popperian epistemology is necessary before artificial general intelligence can be realized.\cite{DeutschAeon} A core part of Popper's philosophy is that he rejected induction and argued that induction is not required for the growth and development of scientific knowledge.\cite{popper1963conjectures} This is in contrast to many present day AI methods which are equivalent to Bayesian methods of induction in how they learn from data.\cite{bishop:2006:PRML,Gal2016DropoutUncertainty} Indeed, an in-vogue view today is that \textit{all} AI systems are an approximation of Solomonoff induction.\cite{HutterAIXI} To explain how science can work without induction Popper argued that theories can only be falsified by evidence, and not confirmed. In Popper's view theories are ``bold conjectures'' invented to solve problems and not learned directly from experience. To give a poignant example, the idea that stars are actually distant suns was a bold conjecture first proposed by Anaxagoras around 450 BC. While how to generate such conjectures is of great interest to AI researchers, Popper was not so much concerned with how conjectures may be generated, seeing that as a question for psychologists  to answer instead.\cite{popper1963conjectures} The truthfulness of a conjecture, after all, is independent of its source. 

While experience can and does inform us as to which conjectures to keep and which to discard, in particular in the form of empirical tests, in Popper's view experience is always theory-laden. In other words, observations cannot be made in an unprejudiced or unbiased manner as Francis Bacon had prescribed.\cite{BaconNovum} To Popper, the question of which comes first, theory or observations, is much like the question of the chicken and the egg.\cite{popper1979objective} Scientific theories are built on observations, which in term were informed by previous scientific theories in a chain going back to pre-scientific myths.\cite{popper1979objective} Thus, while empirical testing of theories plays a role in the form of falsifying certain theories while preserving others, Popper believes that fundamentally all theories originate ``from within'' rather being impressed into the mind from outside.\cite{popper1996myth} Interestingly, Popper conjectured that critical rationalism was not limited to explaining the functioning of science but could be extended to explain elements of psychology as well (ie. how the human brain works and obtains knowledge).\cite{popper1996myth}

If falsifiability is all that matters for a theory to be scientific, should theories based on myths that make falsifiable predictions about the actions of Gods, demons, and ghosts be considered scientific? The need to remove such obviously unscientific theories from the realm of the scientific was a partially addressed by Popper through his notion of ``degrees of falsifiability'', but the extent to which this principle succeeds at the task is unclear.\cite{Popper:LoSD} Deutsch seems to have introduced his HTV principle as an alternative criteria for demarcating scientific vs unscientific theories. The HTV principle asserts that we should reject explanations based on myths because they are easy to vary. Thus, if the data were different, they could easily adapt to that situation. An illustrative example which Deutsch elaborates in detail in his book \textit{The Beginning of Infinity}\cite{Deutsch2011-DEUTBO} and in a 2009 TED talk\cite{DeutschTED} is the ancient Greek myth that the seasons were due to the periodic sadness of the god Demeter. This myth can be easily varied by changing the gods involved and their behaviours, and thus is not a good explanation. The theory can also be varied both without changing the predictions it makes, for instance by substituting the gods employed with other ones, and can also be varied in such a way as to make the predictions adapt to any scenario. Both of these types of variability make it a bad explanation, according to Deutsch.\cite{Deutsch2011-DEUTBO} 

What makes a theory hard-to-vary? For Deutsch a key factor is constraints arising from the internal deductive logic of the theory. As he explains,\cite{Deutsch2011-DEUTBO} the modern day explanation of the seasons is a good explanation because it involves a tight series of geometrical deductions regarding the suns rays and the Earth's axis tilt. There are a few free parameters, such as the angle of tilt, but most of the explanation is rooted in geometrical deductions which cannot be varied. Another important source of constraint is consistency with established knowledge.  

\section{Relation to the bias-variance trade-off and Occam's razor}
So far, the HTV principle has not been formalized. In this section we explore the relation of the HTV principle to two principles which have been formalized - the bias-variance trade-off and Occam's razor. An important point is that while the HTV principle can be applied to particular explanations, it requires the notion that an explanation is a member of a certain equivalence class of possible explanations that it can be varied into. This class must be restricted otherwise every explanation can be varied into any other (with enough possible additions / subtractions) and all explanations are equally variable. In the context of machine learning the HTV principle may be applied, for example, separately to various types of model architecture, each with a fixed number of parameters. Models which have more parameters are capable of expressing a larger class of functions and are therefore easier-to-vary. Superficially therefore the HTV principle seems related to the bias-variance trade-off in classical statistics, which says that models with too many parameters are more prone to overfitting their training data, resulting in poor generalization to test data. However, recall we are interested in extrapolation, not classical generalization within the scope of the training distribution. The bias-variance trade-off only addresses in-distribution generalization, and thus is quite seperate from the HTV principle. Another point is that the bias-variance trade-off has been shown to break down in machine learning as more parameters are added to the model - leading to the ``double descent'' curve, where beyond a certain threshold more parameters always helps and never hurts.\cite{Belkin2019,nakkiran2019deep,Elton2020} So, the bias-variance trade-off is also questionable on its own right. What separates models where the bias-variance trade-off applies from models where it does not is still an ongoing area of research. A final point is that the issue of over-fitting can typically be compensated for with a larger dataset. So large models are not intrinsically bad, as is sometimes assumed from the bias-variance trade-off -- it depends on the quantity of data available. 

The HTV principle appears to be more closely related to Occam's razor, a principle which was stated by William of Occam in the 14th century as ``it is pointless to do with more what can be done with fewer''\cite{summalogicae} and also as ``Plurality (of entities) should not be posited without necessity''.\cite{1495quaestiones} Occam's razor is already deeply embedded in the culture of science and is also quite popular amongst AI researchers. It is discussed in the most popular AI textbooks and the minimum description length principle\cite{Rissanen1978} is an oft-cited formalization of Occam's razor which has been applied in many areas of AI such as regularizing neural networks.\cite{Hinton1993} Solomonoff's theory of universal induction includes a constraint for Occam's razor,\cite{Rathmanner2011} formalized using Kolmogorov complexity as a measure of simplicity. Solomonoff's theory of induction was later used by Hutter to develop the AIXI model of universal artificial intelligence which has been influential in certain quarters of AI research.\cite{HutterAIXI}

There are two ways to measure simplicity. One is to look at any particular trained model or explanatory framework and how complex it is. Hochreiter \& Schmidhuber have found that deep learning models which inhabit flat minima in the loss function surface in parameter space generalize more easily.\cite{Hochreiter1997} The intuition explaining this is that flat-minima indicate a lower-complexity model (more easily compressible). However, counter examples to the idea that deep learning models with lower curvature in their loss function have greater generalizability have been found recently.\cite{DinhSharpMinima} To tie into the work of Popper one can look directly at the function itself, rather than the loss function to measure the complexity of a model. The idea that functions with lower curvature are ``simpler'' than higher curvature functions has a long history and has been criticised by Popper.\cite{Popper:LoSD} To Popper, the practice of ranking functions by their simplicity is done for aesthetic or practical considerations and is not well-grounded in any deep epistemic principle.\cite{Popper:LoSD} 

The other way to measure simplicity is to look at the number of free parameters in a model. This measure appears to align more with Deutsch's concept of variability. Models with more free parameters (or in Occam's language ``entities'') are more variable since the parameters can easily be tweaked to fit different data. Deutsch rejects the converse though - that models with fewer entities are less variable, saying that ``there are plenty of simple explanations that are highly variable, such as `Demeter did it' ''.\cite{Deutsch2011-DEUTBO} It seems what Deutsch is after is looking at the number of constraints on the theory, both internal and external, and perhaps also the degree to which small changes in parameters change the prediction of the model. François Chollet argues that Occam's razor is antithetical to extrapolation.\cite{chollet2019measure} If a model is the simplest possible to achieve good performance on training data, for instance, it is unlikely to do well in new situations. Chollet argues that models should contain ``extraneous''  information in order to be able to extrapolate.  

\section{Measuring internal variability with size of the Rashomon set}
We define internal variability as how much a model/theory can be varied internally while still yielding the same predictions. This is in contrast to external variability, which is how much a model must be varied to fit to new data. In the first case lower variability is better and in the second lower need for variation is better. Let us consider the common problem of predicting a response variables $y$ from input variables $x$ and how it would be approached by both a scientist and a machine learning practitioner. The scientist would first bring in prior knowledge and often, prior known scientific laws that are relevant to the system at hand. Then, using this prior knowledge they would either derive or guess a functional form for the relation, fit it to the data, and see how well it works. It is important the function remains relatively simple, so it remains understandable - there is an implicit desire to understand in addition to accurately predict.  In machine learning, by contrast, the practitioner is only concerned with predictive accuracy, so large black box functions are acceptable. Leo Brieman notes an odd fact that arises when large multi-parameter functions (such as neural nets) are utilized  -- a large multiplicity of models can have equal error (loss) on any dataset $\lbrace x, y \rbrace$. This is true whenever the data is noisy, as is the case in any real-world application. He calls this the Rashomon effect, after a Japanese movie where four people all witness an incident where one person ends up dead. In court, they all report seeing the same facts, but their explanations differ wildly as to what happened. This exact phenomena is observed in neural networks - it is an underappreciated fact that deep neural networks trained with different random initializations may achieve equivalent accuracy but work differently internally (use different features, for instance).\cite{Elton2020} Measuring the size of the Rashomon set of a deep learning model corresponds to determining the number of equivalent minima. This is a level set problem, and as far as we know not much research has been done in this area and there is no easy way of computing the size of this set.  

\section{Measuring external variability using  algorithmic information theory}
Let us assume again we are modeling a simple function $y = f(x)$ and trying to fit it to a dataset $\lbrace (x, y) \rbrace$. To measure external variability, intuitively we wish to know how much we have to change $f(x)$ to adapt the model to changes in the dataset. We can get further intuition about this by considering two radically different types of predictive model - $k$-nearest neighbors and a physics based simulation to compute maximum pressure in a core-collapse supernovae as a function of it's mass. The first model can flexibly fit any function $y = f(x)$ while the second is tailored for a very specific use. Another example of a highly flexible model is a neural network bundled with an optimizer (to make a prediction given the dataset, the neural network is first fit to the data). 

To make this quantitative we must have a way of quantifying the change in the model required to adapt to the new dataset. Algorithmic information theory can help with both these problems. Given a Turing machine and a dataset $D_1$, the algorithmic information in $D_1$, also called Kolmogorov complexity, is the length of bit string $s_1$ which is the shortest program which reproduces $D_1$ using the particular Turing machine we have chosen. It is denoted as $H(D_1) = \hbox{length}(s_1)$. Now suppose we have an AI algorithm $s_{\ff{AI}}$ trained on a different dataset $D_{\ff{AI}}$ and we wish to measure how much it would have to be varied to work optimally on $D_2$. The relative algorithmic information between $s_1$ and $s_{\ff{AI}}$, denoted $H(s_1|s_{\ff{AI}})$ is the length of the shortest algorithm that given $s_{\ff{AI}}$ reproduces $s_1$. Thus, it is a measure of the amount of variation required to make $s_{\ff{AI}}$ work optimally on $D_2$. Unfortunately, Kolmogorov complexity is incomputable. However, we can imagine approximations of this general idea -- instead of trying to find the optimal program for reproducing $D_2$ we specify a margin of accuracy (lossy compression). Then, search to find the minimum change in the length of $s_{\ff{AI}}$ needed to reproduce $D_2$ within the margin we have set. 

Note that we are not interested in how big $s_{\ff{AI}}$ is, only in how much it needs to be varied to fit new incoming data streams. Thus, this notion of HTV-ness is different than Occam's razor. We have still not specified how to set $D_{\ff{AI}}$ (the initial dataset the AI was designed for) and $D_2$. Clearly these datasets cannot be randomly generated. The no-free-lunch theorem states that all algorithms are equivalent when averaged over every possible problem.\cite{Wolpert1997} Thus all algorithms will be on an equal footing when trying to extrapolate from one randomly chosen dataset to another. The situation we are in is very similar to the situation faced by Chollet when trying to rigorously define intelligence.\cite{chollet2019measure}  Chollet's solution was to restrict problem space to a set of problems which humans can solve. Chollet notes that the natural world conforms to a set of very basic priors such as objectness (there exist distinct objects), elementary physics, agentness (that there exist agents with goals), numbers and arithmetic, and elementary geometry and topology. We can similarly assert that the generation of $D_1$ and $D_2$ must conform to a set of prior rules. Or, even simpler, we can state that $D_1$ and $D_2$ are generated by physical phenomena. This limits problem/data space sufficiently that extrapolation becomes feasible.

 
\section{What can we learn from the brain?}
This section may appear to be a lengthy digression, but we believe it is helpful to study the best solution to the intelligence problem that evolution has discovered - the human brain. We distinguish two types of process that may occur in the brain. The first is the varying of randomly-initialized easy-to-vary neuronal models with many parameters to fit incoming sensory data, very similar to what deep learning does. The second process is Popperian in nature and involves the generation of hard-to-vary explanations which must be thrown out ``whole cloth'' if falsified by experience. Distinguishing the relative importance of these two processes for human intelligence has import for futurists and AI safety researchers looking to predict the development trajectory of AI and how long we have until superintelligence is developed.\cite{bostrom2014superintelligence} The differing views on the relative importance of these two types of processes can explain a lot of the variance between Deutsch and other experts regarding the predicted date superintelligent AI will be developed.\cite{Armstrong2015} If the brain works entirely by fitting easy-to-vary models to big data, then reaching human-level intelligence will be possible by scaling up deep learning with larger and larger models and datasets, suggesting a sooner time-to-arrival for superintelligent AI. However, if hard-to-vary explanations are an important part of human intelligence, the problem of how to program an AI to generate/discover such explanations must be solved first before truly human-like AI can be produced. Of course, just as the way airplanes fly is much different from birds, superintelligent general-purpose AI might operate in a very different way from humans and this scenario should not be discounted. In passing we note that Christopher Olah of OpenAI has posited something similar -- in an interview published online he speculates that as deep learning models are scaled up they will eventually reach a point where they start to generate ``crisp abstractions''.\cite{OlahLW} Given that currently many of the abstractions (features) being learned by today's overparametrized deep learning systems are very ``uncrisp'' and brittle,\cite{DBLP:conf/nips/IlyasSTETM19} we find this to be a questionable theory. 

The brain has about 8.6x$10^{10}$ neurons and an average of $10^{3}$ connections (synapses) per neuron. Each synapse contains at least 4.7 bits of information.\cite{BartolBits} Information is encoded in the brain both in the synaptic weights and the pattern of connections, which is arguably much more important.\cite{seung2012connectome} About 80\% of those neurons are in the cerebellum, which is responsible for motor control with almost all the others in the cerebral cortex. Which neurons are connected to which is determined by a mix of genetic mechanisms which control the placement of connections during development and processes of dendritic growth and pruning during the lifetime of any given individual in response to its environment. The former can be viewed as an ``outer-loop'' optimization that occurs via evolution over many generations while the latter is part of an ``inner-loop'' optimization that occurs during an individual's lifetime. As a side note, Popper has noted that evolution, since it is based on random mutations which happen ``from within'' operates along the lines of his epistemology (conjecture and criticism/falsification).\cite{popper1996myth} The ``inner loop'' optimization, by contrast, appears to be largely based on fitting sense data. Both processes are known to occur although the relative importance of each is far from understood (the nature-vs-nurture debate).\cite{Zador2019} 
We conjecture that to the extent the HTV principle is involved in human cognition it would have to have arisen from genetics, having been stumbled on by evolution and not learned by fitting highly variable neural models to sense data. The amount of data that can inform neural structure from genetics is limited however by size of the genome. This ``genetic bottleneck'' has been suggested to serve as a form of regularization on how evolution has shaped the brain.\cite{Zador2019} The human genome contains about 750 megabytes of information total. About 2\% of the genome is coding for proteins (15 Mb). While the degree of importance of the non-coding genome is largely unknown, if we assume 25\% of the non-coding genome is used for regulatory purposes, that is about 200 Mb for the genetic bottleneck. 

In what follows, we ignore the complex question of the importance of how much is learned from genetics vs environment. Instead we focus on the question of how far brute-force fitting to sense data alone can go. Conventional thinking in neuroscience has been that the brain cannot operate on brute-force fitting of sense data alone, because the amount of data provided is too scant to train the number of parameters involved. As an aside, this ``poverty of the stimulus'' was the key motivation for Chomsky to propose a genetic component to human grammar which he called ``universal grammar''.\cite{chomsky2014aspects} However, recently Hasson et al. have argued that the amount of data streaming in though the senses is enough to allow for brute-force fitting.\cite{Hasson2019} Quantifying the amount of information streaming into the brain through the senses is very difficult, but it may be quite large -- Zimmerman estimates that the channel capacity of the human visual system is 10$^7$ bits/s while the channel capacity of the auditory system is 10$^5$ bits/s.\cite{Zimmermann1986} The actual amount of information is much lower due to redundancy effects, however if we assume it is 100 times slower than learning the 4x$10^{14}$ bits necessary to specify all 9x$10^{13}$ synapses in the brain is theoretically possible in 133 years. Regardless of the relevance of this back-of-the envelope calculation, it is safe to say that the brain does process very large amounts of data. For instance a child has seen many faces by an early age under a variety of angles, distances, and lighting conditions. As Hasson points out, humans are subject to an ``other-race'' effect whereby they find faces of different ethnicities more difficult to distinguish, suggesting a type of brute force fitting which struggles to extrapolate even slightly.\cite{Hasson2019} A study based on audio recording in a child's home suggests that children hear at least 5-10 million words per year.\cite{Roy2015} By age 10, this suggests 100 million words may have been heard. This is very large but still 3000 times smaller than the natural language model GPT3's training corpus, which contained 300 billion tokens.\cite{brown2020language}  
\begin{figure}[ht]
    \includegraphics[width=\textwidth]{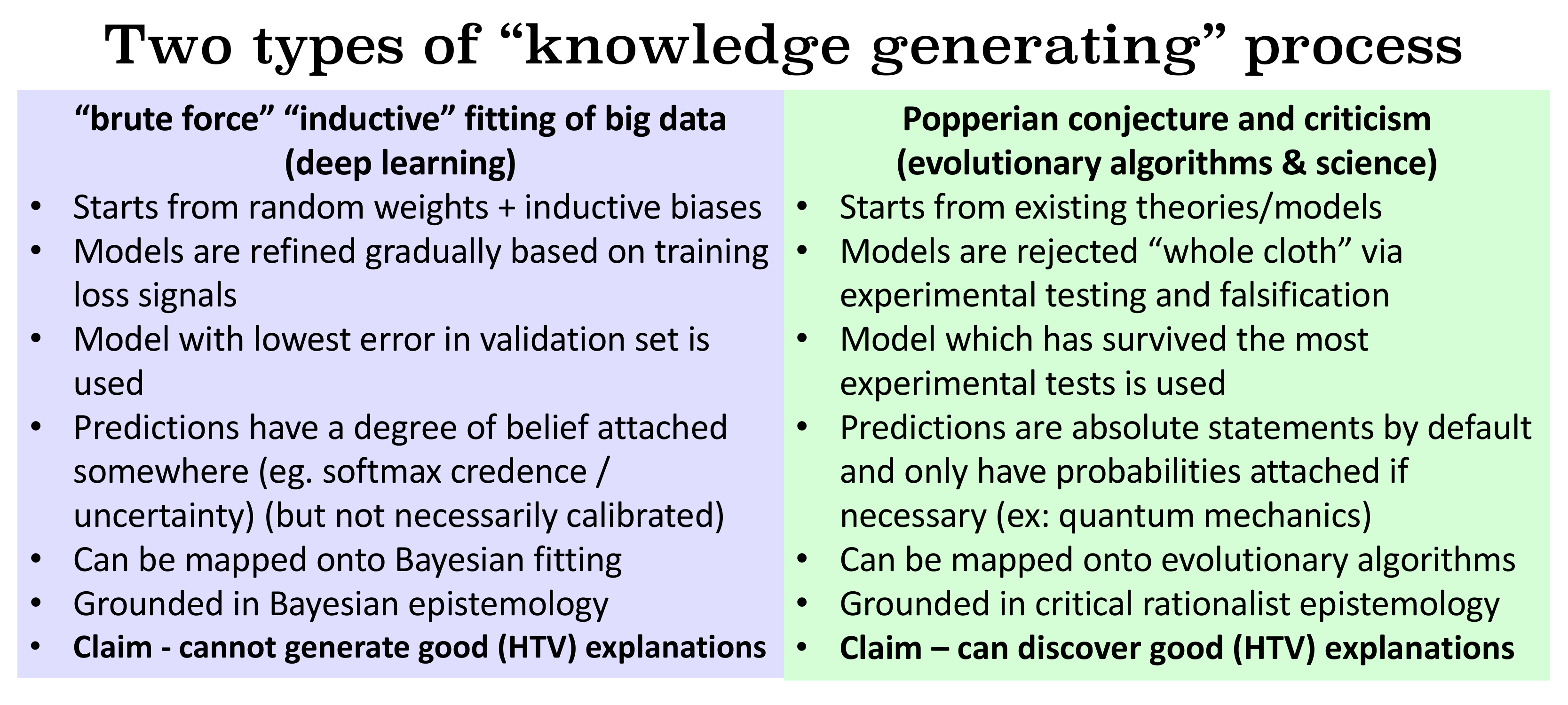}
    \caption{Summary of two types of process which occur in the brain and seem to map onto Kahneman's System I and System II.} 
    \label{fig:diagram}
\end{figure}

Levy, Hasson, and Malach have estimated that two hundred million neurons in the visual cortex are involved in the representation of a single-object image.\cite{Levy2004} Thus, it seems that a large part of the brain, at least the part dedicated to raw perception of objects and possibly a large part of language ability, is learned by fitting large easy-to-vary neural models to sensory data. However even Uri Hasson, who forcefully argued for this position recently,\cite{Hasson2019} points out that humans have additional abilities which go beyond this, pointing to our ability to learn to mathematics such as calculus and perform abstract symbolic reasoning.\cite{PolichHasson} 

What about ``common sense'' knowledge, or ``background knowledge''? Some common sense knowledge comes in the form of ``rules of thumb'', which are rules extracted from experience through induction. They are brute facts, often ``free floating'' and unconnected to other facts. Suppose we are talking to someone who has no knowledge of how the human body works. They still may be able to treat certain diseases using a long list of rules of thumb gained from experience. For instance:  
\begin{enumerate}
    \item Aspirin helps with headaches. If Aspirin doesn't work, try ibuprofen next. 
    \item Drinking lots of water helps ameliorate cold symptoms. 
    \item Homeopathic remedies do not work. (note one can mistakenly go the wrong way here!) 
\end{enumerate}
The person wouldn't be able to explain any of these facts but instead would justify them by pointing to how often they have been confirmed by prior experience. Explanatory theories, on the other hand, are required whenever trying to predict something you haven't seen before. Consider the following questions: 
\begin{itemize}
    \item What would happen if you poured a bottle of bleach into the fuel tank of a car? 
    \item What would happen if the United States cut off all trade with France tomorrow? 
    \item What would happen if you diluted hot sauce 10 times and repeated this 10x dilution process 10 times?
\end{itemize}

We must use some form of common sense reasoning to answer these questions, reasoning which uses explanatory theories. The GPT2 language model (which is a type of brute-force fit, highly-variable model) is very bad at answering these sorts of questions\cite{marcus2020gpt2}. GPT3 does much better, but still stumbles in this particular area.\cite{VanceGPT3}

The two processes discussed in the last section map nicely onto Kahneman's distinction between System 1 and System 2.\cite{kahneman2011thinking} Some unique characteristics of both of these two processes are summarized in figure \ref{fig:diagram}. To summarize our view on the titular question of this section, it appears that large parts of the brain such as the perceptual and motor systems are learned by brute-force fitting. However, our capacity to build both common sense explanatory theories and scientific explanatory theories is where the HTV principle seems to be important. 

\section{Conclusion}
Present day AI is largely built on top of two philosophical principles - induction and Occam's razor. Popper and Deutsch reject induction, arguing for critical rationalism instead, which is based on conjecture, criticism, and falsification. Deutsch also rejects Occam's razor, calling it a ``misconception''.\cite{Deutsch2011-DEUTBO} The HTV principle was introduced by Deutsch to improve critical rationalist epistemology. However, the HTV principle may be a useful principle to guide further AI research regardless of which view on epistemology one deems as correct. In this preliminary work we explored the relevance of this principle to artificial intelligence and neuroscience. The brain appears to operate largely on brute-force fitting of easy-to-vary models. At the same time however, humans possess additional capacities for reasoning (ie Kahneman's  ``system 2'') and are capable of inventing explanations (both of the common sense and scientific variety) which are good explanations capable of extrapolation. Both of these seem impossible to obtain with brute-force fitting of highly variable models alone. The HTV principle is very similar to Occam's razor, but distinct if one accepts that simple theories can be highly variable. Additionally, the HTV principle deals not with model simplicity (as measured for instance in an information theoretic way), but with the number of ways a model can be varied and how ``hard'' the variation is (ie how much has to be changed to adapt to new situations or training data). We presented some tentative methods for measuring how hard to vary a model/explanation is. We believe work in this area may shed light on a key issue facing AI research today - how to build systems that can generate ``good explanations'' of the world which allows them to function in new situations. As François Chollet has beautifully argued recently, the true test of intelligence is how well a system can adapt to solve new environments and tackle entirely new problem situations.\cite{chollet2019measure}

\subsection*{Funding \& disclaimer}
No funding sources were used in the creation of this work. The author wrote this article in his personal capacity. The opinions expressed in this article are the author's own and do not reflect the view of the National Institutes of Health, the Department of Health and Human Services, or the United States government.

\subsection*{Acknowledgements}
The author appreciates helpful discussions with Dr.\ Felix Faber on this subject and feedback from Bruce Nielson, who has written several blog posts on the HTV principle and critical rationalism on Medium and his blog (\href{http://fourstrands.org/}{http://fourstrands.org/}). The author acknowledges helpful exchanges with Dennis C. Hackethal and Ella Hoeppner, both of whom have written about Popperian epistemology and AI.

%
\bibliographystyle{elsarticle-num}
\bibliography{bibliography.bib}

\begin{thebibliography}{10}
\expandafter\ifx\csname url\endcsname\relax
  \def\url#1{\texttt{#1}}\fi
\expandafter\ifx\csname urlprefix\endcsname\relax\def\urlprefix{URL }\fi
\expandafter\ifx\csname href\endcsname\relax
  \def\href#1#2{#2} \def\path#1{#1}\fi

\bibitem{He2015}
K.~He, X.~Zhang, S.~Ren, J.~Sun, Delving deep into rectifiers: Surpassing
  human-level performance on {ImageNet} classification, in: 2015 {IEEE}
  International Conference on Computer Vision ({ICCV}), {IEEE}, 2015, p. 1026.

\bibitem{Liu2019}
X.~Liu, L.~Faes, A.~U. Kale, S.~K. Wagner, D.~J. Fu, A.~Bruynseels,
  T.~Mahendiran, G.~Moraes, M.~Shamdas, C.~Kern, J.~R. Ledsam, M.~K. Schmid,
  K.~Balaskas, E.~J. Topol, L.~M. Bachmann, P.~A. Keane, A.~K. Denniston, A
  comparison of deep learning performance against health-care professionals in
  detecting diseases from medical imaging: a systematic review and
  meta-analysis, The Lancet Digital Health 1~(6) (2019) e271--e297.

\bibitem{Silver2016}
D.~Silver, A.~Huang, C.~J. Maddison, A.~Guez, L.~Sifre, G.~van~den Driessche,
  J.~Schrittwieser, I.~Antonoglou, V.~Panneershelvam, M.~Lanctot, S.~Dieleman,
  D.~Grewe, J.~Nham, N.~Kalchbrenner, I.~Sutskever, T.~Lillicrap, M.~Leach,
  K.~Kavukcuoglu, T.~Graepel, D.~Hassabis, Mastering the game of {Go} with deep
  neural networks and tree search, Nature 529~(7587) (2016) 484--489.

\bibitem{Deutsch2011-DEUTBO}
D.~Deutsch, The Beginning of Infinity: Explanations That Transform the World,
  Viking Adult, 2011.

\bibitem{hastie01statisticallearning}
T.~Hastie, R.~Tibshirani, J.~Friedman, The Elements of Statistical Learning,
  Springer Series in Statistics, Springer New York Inc., New York, NY, USA,
  2001.

\bibitem{NeyshaburICLR2020}
Y.~Jiang, B.~Neyshabur, H.~Mobahi, D.~Krishnan, S.~Bengio, Fantastic
  generalization measures and where to find them, in: 8th International
  Conference on Learning Representations, {ICLR} 2020, Addis Ababa, Ethiopia,
  April 26-30, 2020, OpenReview.net, 2020.

\bibitem{Beede2020}
E.~Beede, E.~Baylor, F.~Hersch, A.~Iurchenko, L.~Wilcox, P.~Ruamviboonsuk,
  L.~M. Vardoulakis, A human-centered evaluation of a deep learning system
  deployed in clinics for the detection of diabetic retinopathy, in:
  Proceedings of the 2020 {CHI} Conference on Human Factors in Computing
  Systems, {ACM}, 2020, pp. 1--12.

\bibitem{simonite_2018}
T.~Simonite,
  \href{https://www.wired.com/story/when-it-comes-to-gorillas-google-photos-remains-blind/}{When
  it comes to gorillas, google photos remains blind}, WIRED.
\newline\urlprefix\url{https://www.wired.com/story/when-it-comes-to-gorillas-google-photos-remains-blind/}

\bibitem{Mnih2015}
V.~Mnih, K.~Kavukcuoglu, D.~Silver, A.~A. Rusu, J.~Veness, M.~G. Bellemare,
  A.~Graves, M.~Riedmiller, A.~K. Fidjeland, G.~Ostrovski, S.~Petersen,
  C.~Beattie, A.~Sadik, I.~Antonoglou, H.~King, D.~Kumaran, D.~Wierstra,
  S.~Legg, D.~Hassabis, Human-level control through deep reinforcement
  learning, Nature 518~(7540) (2015) 529--533.

\bibitem{Kansky2017}
K.~Kansky, T.~Silver, D.~A. M{\'{e}}ly, M.~Eldawy, M.~L{\'{a}}zaro{-}Gredilla,
  X.~Lou, N.~Dorfman, S.~Sidor, D.~S. Phoenix, D.~George, Schema networks:
  Zero-shot transfer with a generative causal model of intuitive physics, in:
  D.~Precup, Y.~W. Teh (Eds.), Proceedings of the 34th International Conference
  on Machine Learning, {ICML} 2017, Sydney, NSW, Australia, 6-11 August 2017,
  Vol.~70 of Proceedings of Machine Learning Research, {PMLR}, 2017, pp.
  1809--1818.

\bibitem{Marcus2018}
G.~Marcus,
  \href{https://www.wired.com/story/deepminds-losses-future-artificial-intelligence/}{Deepmind's
  losses and the future of artificial intelligence}, WIRED.
\newline\urlprefix\url{https://www.wired.com/story/deepminds-losses-future-artificial-intelligence/}

\bibitem{Popper:LoSD}
K.~R. Popper, The Logic of Scientific Discovery, Hutchinson, London, 1934.

\bibitem{Hasson2019}
U.~Hasson, S.~A. Nastase, A.~Goldstein, Direct fit to nature: An evolutionary
  perspective on biological and artificial neural networks, Neuron 105~(3)
  (2020) 416--434.

\bibitem{Elton2020}
D.~C. Elton, Self-explaining {AI} as an alternative to interpretable {AI}, in:
  Artificial General Intelligence, Springer International Publishing, 2020, pp.
  95--106.

\bibitem{sucholutsky2020less}
I.~Sucholutsky, M.~Schonlau, `less than one'-shot learning: Learning n classes
  from m$<$n samples, arXiv e-prints: 2009.08449.

\bibitem{1920starlightbent}
F.~W. Dyson, A.~S. Eddington, C.~Davidson, A determination of the deflection of
  light by the sun's gravitational field, from observations made at the total
  eclipse of {May} 29, 1919, Philosophical Transactions of the Royal Society of
  London. Series A, Containing Papers of a Mathematical or Physical Character
  220~(571-581) (1920) 291--333.

\bibitem{Elton2018}
D.~C. Elton, Z.~Boukouvalas, M.~S. Butrico, M.~D. Fuge, P.~W. Chung, Applying
  machine learning techniques to predict the properties of energetic materials,
  Scientific Reports 8~(1).

\bibitem{barnes2018machine}
B.~C. Barnes, D.~C. Elton, Z.~Boukouvalas, D.~E. Taylor, W.~D. Mattson, M.~D.
  Fuge, P.~W. Chung, Machine learning of energetic material properties, arXiv
  eprints: 1807.06156.

\bibitem{Zhavoronkov2019}
A.~Zhavoronkov, Y.~A. Ivanenkov, A.~Aliper, M.~S. Veselov, V.~A. Aladinskiy,
  A.~V. Aladinskaya, V.~A. Terentiev, D.~A. Polykovskiy, M.~D. Kuznetsov,
  A.~Asadulaev, Y.~Volkov, A.~Zholus, R.~R. Shayakhmetov, A.~Zhebrak, L.~I.
  Minaeva, B.~A. Zagribelnyy, L.~H. Lee, R.~Soll, D.~Madge, L.~Xing, T.~Guo,
  A.~Aspuru-Guzik, Deep learning enables rapid identification of potent {DDR}1
  kinase inhibitors, Nature Biotechnology 37~(9) (2019) 1038--1040.

\bibitem{Walters2020}
W.~P. Walters, M.~Murcko, Assessing the impact of generative {AI} on medicinal
  chemistry, Nature Biotechnology 38~(2) (2020) 143--145.

\bibitem{Elton2019}
D.~C. Elton, Z.~Boukouvalas, M.~D. Fuge, P.~W. Chung, Deep learning for
  molecular design{\textemdash}a review of the state of the art, Molecular
  Systems Design {\&} Engineering 4~(4) (2019) 828--849.

\bibitem{DeutschAeon}
D.~Deutsch,
  \href{https://aeon.co/essays/how-close-are-we-to-creating-artificial-intelligence}{Creative
  blocks}, Aeon.
\newline\urlprefix\url{https://aeon.co/essays/how-close-are-we-to-creating-artificial-intelligence}

\bibitem{popper1963conjectures}
K.~Popper, Conjectures and Refutations: The Growth of Scientific Knowledge,
  Routledge \& K. Paul, 1963.

\bibitem{bishop:2006:PRML}
C.~M. Bishop, Pattern Recognition and Machine Learning, Springer, 2006.

\bibitem{Gal2016DropoutUncertainty}
Y.~Gal, Z.~Ghahramani, Dropout as a {Bayesian} approximation: Representing
  model uncertainty in deep learning, in: M.~F. Balcan, K.~Q. Weinberger
  (Eds.), Proceedings of The 33rd International Conference on Machine Learning,
  Vol.~48 of Proceedings of Machine Learning Research, PMLR, New York, New
  York, USA, 2016, pp. 1050--1059.

\bibitem{HutterAIXI}
M.~Hutter, A theory of universal artificial intelligence based on algorithmic
  complexity, arXiv e-prints: cs/0004001.

\bibitem{BaconNovum}
F.~Bacon, Novum Organum, 1620.

\bibitem{popper1979objective}
K.~Popper, Objective Knowledge: An Evolutionary Approach, Clarendon Press,
  1979.

\bibitem{popper1996myth}
K.~Popper, K.~Popper, M.~Notturno, The Myth of the Framework: In Defence of
  Science and Rationality, In Defence of Science and Rationality, Routledge,
  1996.

\bibitem{DeutschTED}
D.~Deutsch,
  \href{https://www.ted.com/talks/david_deutsch_a_new_way_to_explain_explanation}{A
  new way to explain explanation}, TED Talk.
\newline\urlprefix\url{https://www.ted.com/talks/david_deutsch_a_new_way_to_explain_explanation}

\bibitem{Belkin2019}
M.~Belkin, D.~Hsu, S.~Ma, S.~Mandal, Reconciling modern machine-learning
  practice and the classical bias{\textendash}variance trade-off, Proceedings
  of the National Academy of Sciences 116~(32) (2019) 15849--15854.

\bibitem{nakkiran2019deep}
P.~Nakkiran, G.~Kaplun, Y.~Bansal, T.~Yang, B.~Barak, I.~Sutskever, Deep double
  descent: Where bigger models and more data hurt, arXiv eprints: 1912.02292.

\bibitem{summalogicae}
W.~of~Ockham, Summa totius logicae, 1323.

\bibitem{1495quaestiones}
J.~Badius, J.~Trechsel,
  \href{https://books.google.com/books?id=1uSplwEACAAJ}{Quaestiones et
  decisiones in quattuor libros Sententiarum Petri Lombardi: Centilogium
  theologicum}, Johannes Trechsel, 1495.
\newline\urlprefix\url{https://books.google.com/books?id=1uSplwEACAAJ}

\bibitem{Rissanen1978}
J.~Rissanen, Modeling by shortest data description, Automatica 14~(5) (1978)
  465--471.

\bibitem{Hinton1993}
G.~E. Hinton, D.~van Camp, Keeping the neural networks simple by minimizing the
  description length of the weights, in: Proceedings of the sixth annual
  conference on Computational learning theory - {COLT} {'}93, {ACM} Press,
  1993.

\bibitem{Rathmanner2011}
S.~Rathmanner, M.~Hutter, A philosophical treatise of universal induction,
  Entropy 13~(6) (2011) 1076--1136.

\bibitem{Hochreiter1997}
S.~Hochreiter, J.~Schmidhuber, Flat minima, Neural Computation 9~(1) (1997)
  1--42.

\bibitem{DinhSharpMinima}
L.~Dinh, R.~Pascanu, S.~Bengio, Y.~Bengio, Sharp minima can generalize for deep
  nets, in: D.~Precup, Y.~W. Teh (Eds.), Proceedings of the 34th International
  Conference on Machine Learning, {ICML} 2017, Sydney, NSW, Australia, 6-11
  August 2017, Vol.~70 of Proceedings of Machine Learning Research, {PMLR},
  2017, pp. 1019--1028.

\bibitem{chollet2019measure}
F.~Chollet, On the measure of intelligence, arXiv e-prints: 1911.01547.

\bibitem{Wolpert1997}
D.~Wolpert, W.~Macready, No free lunch theorems for optimization, {IEEE}
  Transactions on Evolutionary Computation 1~(1) (1997) 67--82.

\bibitem{bostrom2014superintelligence}
N.~Bostrom,
  \href{https://books.google.com/books?id=7\_H8AwAAQBAJ}{Superintelligence:
  Paths, Dangers, Strategies}, Oxford University Press, 2014.
\newline\urlprefix\url{https://books.google.com/books?id=7\_H8AwAAQBAJ}

\bibitem{Armstrong2015}
S.~Armstrong, K.~Sotala, How we're predicting {AI} {\textendash} or failing to,
  in: Topics in Intelligent Engineering and Informatics, Springer International
  Publishing, 2015, pp. 11--29.

\bibitem{OlahLW}
E.~Hubinger, {Chris Olah’s views on AGI safety},
  \textit{\url{https://www.lesswrong.com/posts/X2i9dQQK3gETCyqh2/chris-olah-s-views-on-agi-safet}}
  (2020).

\bibitem{DBLP:conf/nips/IlyasSTETM19}
A.~Ilyas, S.~Santurkar, D.~Tsipras, L.~Engstrom, B.~Tran, A.~Madry, Adversarial
  examples are not bugs, they are features, in: H.~M. Wallach, H.~Larochelle,
  A.~Beygelzimer, F.~d'Alch{\'{e}}{-}Buc, E.~B. Fox, R.~Garnett (Eds.),
  Advances in Neural Information Processing Systems 32: Annual Conference on
  Neural Information Processing Systems 2019, NeurIPS 2019, 8-14 December 2019,
  Vancouver, BC, Canada, 2019, pp. 125--136.

\bibitem{BartolBits}
T.~M. Bartol, C.~Bromer, J.~Kinney, M.~A. Chirillo, J.~N. Bourne, K.~M. Harris,
  T.~J. Sejnowski, Nanoconnectomic upper bound on the variability of synaptic
  plasticity, {eLife}.

\bibitem{seung2012connectome}
S.~Seung, \href{https://books.google.com/books?id=GXwEuoYl3wQC}{Connectome: How
  the Brain's Wiring Makes Us Who We Are}, Houghton Mifflin Harcourt, 2012.
\newline\urlprefix\url{https://books.google.com/books?id=GXwEuoYl3wQC}

\bibitem{Zador2019}
A.~M. Zador, A critique of pure learning and what artificial neural networks
  can learn from animal brains, Nature Communications 10~(1).

\bibitem{chomsky2014aspects}
N.~Chomsky, \href{https://books.google.com/books?id=ljFkBgAAQBAJ}{Aspects of
  the Theory of Syntax}, Aspects of the Theory of Syntax, MIT Press, 2014.
\newline\urlprefix\url{https://books.google.com/books?id=ljFkBgAAQBAJ}

\bibitem{Zimmermann1986}
M.~Zimmermann, Neurophysiology of sensory systems, in: Fundamentals of Sensory
  Physiology, Springer Berlin Heidelberg, 1986, pp. 68--116.

\bibitem{Roy2015}
B.~C. Roy, M.~C. Frank, P.~DeCamp, M.~Miller, D.~Roy, Predicting the birth of a
  spoken word, Proceedings of the National Academy of Sciences 112~(41) (2015)
  12663--12668.

\bibitem{brown2020language}
T.~B. Brown, B.~Mann, N.~Ryder, M.~Subbiah, J.~Kaplan, P.~Dhariwal,
  A.~Neelakantan, P.~Shyam, G.~Sastry, A.~Askell, S.~Agarwal, A.~Herbert-Voss,
  G.~Krueger, T.~Henighan, R.~Child, A.~Ramesh, D.~M. Ziegler, J.~Wu,
  C.~Winter, C.~Hesse, M.~Chen, E.~Sigler, M.~Litwin, S.~Gray, B.~Chess,
  J.~Clark, C.~Berner, S.~McCandlish, A.~Radford, I.~Sutskever, D.~Amodei,
  Language models are few-shot learners, arXiv e-prints: 2005.14165.

\bibitem{Levy2004}
I.~Levy, U.~Hasson, R.~Malach, One picture is worth at least a million neurons,
  Current Biology 14~(11) (2004) 996--1001.

\bibitem{PolichHasson}
K.~Polich, ``{Robust Fit to Nature}'' - interview with {Uri Hasson},
  \textit{\url{https://dataskeptic.com/blog/episodes/2020/robust-fit-to-nature}}
  (2020).

\bibitem{marcus2020gpt2}
G.~Marcus, {GPT}-2 and the nature of intelligence, The Gradient.

\bibitem{VanceGPT3}
A.~Vance, Fun with {GPT-3},
  \textit{\url{https://rationalconspiracy.com/2020/07/31/fun-with-gpt-3/}}
  (2020).

\bibitem{kahneman2011thinking}
D.~Kahneman, Thinking, fast and slow, Farrar, Straus and Giroux, New York,
  2011.

\end{thebibliography}

\end{document}